\documentclass[conference]{IEEEtran}

\IEEEoverridecommandlockouts

\usepackage[T1]{fontenc}
\usepackage[utf8]{inputenc}
\usepackage{cite}
\usepackage{graphicx}
\usepackage{booktabs}
\usepackage{float}
\usepackage{amsmath}
\usepackage{url}
\usepackage{microtype}
\usepackage{inconsolata}
\usepackage[hidelinks]{hyperref}

\AddToHook{shipout/foreground}{
  \put(48,-735){\rule{516pt}{0.35pt}}
  \put(48,-748){\makebox(0,0)[bl]{\fontsize{7}{8}\selectfont \textcopyright\ 2026 IEEE. This manuscript has been accepted for publication in the 2026 IEEE 12th International Conference on Collaboration and Internet Computing (CIC).}}
  \ifnum\value{page}=1
    \put(48,-776){\parbox[b]{516pt}{\raggedright\fontsize{6}{7}\selectfont \textcopyright\ 2026 IEEE. Personal use of this material is permitted. Permission from IEEE must be obtained for all other uses, in any current or future media, including reprinting/republishing this material for advertising or promotional purposes, creating new collective works, for resale or redistribution to servers or lists, or reuse of any copyrighted component of this work in other works.}}
  \fi
}

\title{Sentence Specificity Scores for Collaborative Technical Documentation: A Domain-Transfer Study}

\author{\IEEEauthorblockN{\begin{tabular}{c@{\hspace{1.5em}}c@{\hspace{1.5em}}c}
Rocker D'Antonio & Thomas Benton Townsend & Dimitrios Michael Manias \\
orcid: 0009-0001-8444-1772 & orcid: 0009-0004-1100-3474 & orcid: 0000-0003-4390-3093
\end{tabular}}
\IEEEauthorblockA{\textit{Mississippi State University}, USA}\thanks{Research supported by the NVIDIA Academic Grant Program using DGX Spark.}}

\begin{document}
\bstctlcite{IEEEtranBSTCTL:camera-ready}
\maketitle

\begin{abstract}
Collaboration depends on shared context, and technical documentation is one way that context persists across people and AI teammates. Specificity, the amount and exactness of detail expressed in language, shapes what information documentation captures and how precisely that information is communicated. This work audits sentence-specificity scoring artifacts on technical documentation and tests whether scores applied only after generation help choose among fixed LLM-generated revisions. Across Wikipedia and three technical-documentation corpora, the fixed general-domain predictor SpeciTeller and the pinned post-publication author-repository implementation of Ko \textit{et al.}'s target-adapted predictor produce different corpus orders and same-sentence rank agreement from $-0.066$ to $0.510$. Strict filtering and token-length adjustment change these patterns without reconciling them. In the Gemma set, SpeciTeller ranking raises direction-valid selection from 71.7\% to 83.3\% (+11.7 points; 95\% source-case bootstrap interval +1.7 to +21.7); in the GPT-OSS-120B set, SpeciTeller ranking raises direction-valid selection from 51.7\% to 56.7\% (+5.0 points; 95\% source-case bootstrap interval $-6.7$ to +16.7), and every primary single-score GPT-OSS-120B interval includes zero. These findings tie score interpretation and decision value to the predictor and candidate set.
\end{abstract}


\section{Introduction}

Collaboration turns separate knowledge into shared capability: ideas and domain expertise move among participants, enabling work that none could accomplish alone. In systems engineering, technical documentation gives that capability a durable form, preserving context beyond an immediate exchange even when authors and readers understand the same document differently \cite{deboer2009documentation}. Human--agent research has likewise shown how shared representations of tasks, roles, goals, and progress can serve as coordination resources \cite{scheutz2017shared}. Recent multi-agent evidence further shows that the organization and reliability of shared context can affect downstream performance: compact, verified records of findings, failures, and constraints improved software-engineering and multi-document reasoning \cite{mao2026delm}. Through documentation, people and AI teammates retain persistent procedural and operational context.

Since useful detail depends on purpose, the documentation challenge is to determine which steps, parameters, constraints, or entities belong in the shared resource. The specificity of that resource matters to both people and AI systems that consume it as context. In prior work, pairs that successfully solved a task used more specific feature combinations to establish a shared perspective \cite{hayashi2019specificity}.

Specificity can be measured at the sentence level, but the meaning of that measurement may depend on domain. General-domain prose and technical documentation express detail through different cues, so identical scores need not represent the same kind of information in both settings. Their interpretation can therefore be evaluated against multiple forms of evidence. Their effects on downstream decisions can also be measured when the scores are used as signals for steering document refinement and prose generation.

This study examines sentence specificity as a property of collaborative technical context and as a signal for a bounded documentation decision. We ask three questions about sentence-specificity scores in technical documentation. \textit{Interpretation:} when scoring artifacts are applied across domains, how do their score distributions, corpus orderings, and same-sentence rankings change after filtering and length adjustment? \textit{Correspondence:} within technical corpora, do the scores respond to controlled edits and agree locally with related measures and reader judgments? \textit{Decision value:} among score-blind generated revisions, does post-generation ranking select a requested specificity direction more often than choosing the first candidate? This sequence keeps corpus-level score differences distinct from reader agreement and decision benefit.

\begin{figure}[t]
\centering
\includegraphics[width=0.94\columnwidth]{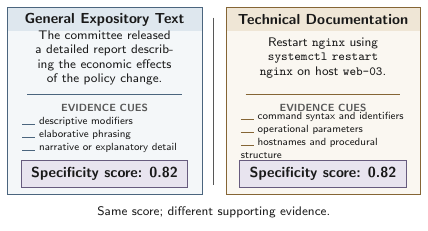}
\caption{Similar specificity scores may be supported by different structural evidence under cross-domain application. The matched score is illustrative, not a verified sentence-level measurement.}
\label{fig:domain_transfer_concept}
\end{figure}

We use \textit{cross-domain evaluation} for applying the scoring procedures described below to a different text domain and assessing their behavior; we do not train a new transfer-learning model. SpeciTeller \cite{li2015speciteller} is applied without adaptation, while the pinned post-publication author-repository implementation of Ko \textit{et al.}'s predictor performs its documented target adaptation in each run. General-domain predictions reflect lexical elaboration, length, and rarity, whereas technical sentences encode commands, identifiers, parameters, steps, and constraints. Figure~\ref{fig:domain_transfer_concept} illustrates how similar scores can therefore rest on different evidence. Domain shift can change model behavior \cite{ben2009theory}, and cross-domain degradation is documented in NLP \cite{zhou2024crossdomain}. The interpretation question is thus not only whether corpus averages differ, but whether each predictor follows the same sentence evidence after filtering and length adjustment.

The decision analysis uses two score-blind candidate sets, each fully rated by one blinded evaluator. For each source, an LLM generates fixed alternatives, policies rank them, and the evaluator rates requested-direction agreement. Guidance on human--AI interaction emphasizes correction and user control \cite{amershi2019guidelines}, motivating a possible author-controlled extension in which a person could decide which ranked alternative would enter documentation.

The results show predictor- and domain-dependent interpretation alongside setting-specific decision value. SpeciTeller and the pinned comparator disagree on corpus order and same-sentence ranking after filtering and length adjustment. In the Gemma set, SpeciTeller selects 83.3\% direction-valid candidates versus 71.7\% for the first candidate, a +11.7-point gain (95\% source-case bootstrap interval +1.7 to +21.7); in the GPT-OSS-120B set, SpeciTeller selects 56.7\% versus 51.7\% for the first candidate, a +5.0-point gain (95\% source-case bootstrap interval $-6.7$ to +16.7), and every primary single-score GPT-OSS-120B interval includes zero.

The contributions of this work are as follows:

\begin{enumerate}
    \item a cross-domain audit of how two specificity predictors score technical sentences;
    \item an evidence structure that separates score interpretation, local correspondence, and value for a named decision, preventing evidence at one stage from substituting for another;
    \item a bounded post-generation test on fixed, score-blind candidates, with one positive Gemma result and unresolved primary single-score gains for GPT-OSS-120B; and
    \item SpecTech, the pinned implementation, provenance, and rerun material for this study.
\end{enumerate}

\section{Terminology}

\textit{Measurement terms.} \textit{Specificity} is the amount and precision of detail conveyed. This study examines specificity at the sentence level, where individual sentences collectively contribute to the context carried by a document. A \textit{specificity predictor} maps a sentence to one \textit{score}; predictors may use different evidence and scales. \textit{Documentation context} is a corpus or domain and its conventions. A score's meaning is the textual evidence it reflects and how it orders sentences in that context.

\textit{Decision terms.} In \textit{candidate selection}, a generator creates fixed alternatives without scores; \textit{candidate admission} applies integrity checks, and \textit{post-generation ranking} then orders admitted alternatives. \textit{Direction-valid} means blinded review finds the revision adds, reduces, or preserves specificity as requested.

This study does not treat these measurements as measures of documentation quality or assess their effects on documentation quality.

\section{Related Work}

\subsection{Specificity and Scalar Sentence Modeling}

Sentence-level specificity prediction was introduced by Li and Nenkova \cite{li2015speciteller}, who formulated specificity as a regression problem and trained SpeciTeller primarily on news-domain data. Subsequent work examined specificity in additional settings, including educational discourse \cite{lugini-litman-2017-predicting}, and proposed domain-agnostic real-valued prediction frameworks \cite{ko2019domain}. These approaches build on earlier notions of lexical specificity in information retrieval, particularly inverse document frequency \cite{SparckJones1972}.

Task-based studies place specificity-related signals in decisions beyond predictor evaluation. Li and Nenkova used predicted specificity to identify simplification targets and characterize original-to-simplified changes \cite{li2015speciteller}. Ko \textit{et al.}\ filtered dialogue training data by specificity rather than length; Remove-General reduced perplexity from 51.97 to 46.92 and was preferred to Remove-Short for informativeness (38.9\% versus 29.0\%; 32.1\% ties) and quality (34.7\% versus 28.9\%; 36.4\% ties) \cite{ko2019domain}. On ClarifyBench, structured clarification for ambiguous or incomplete tool instructions achieved 7--39\% higher coverage with 1.5--2.7$\times$ fewer questions than strong prompting and uncertainty-based baselines \cite{suri-etal-2026-structured}.

GranuScore places text on a continuum from fine-grained entities to broader descriptions \cite{ellinger2026granuscore}. Granularity concerns the abstraction level of referents, whereas specificity concerns the overall amount of concrete detail. Replacing a broad referent such as \textit{software} with \texttt{nginx} can make a sentence both finer-grained and more specific. By contrast, adding when or why to restart \texttt{nginx} can increase specificity without changing the abstraction level of the referent. We therefore use GranuScore as a related comparison rather than as a third specificity predictor.

Related scalar modeling includes sentiment polarity and intensity \cite{tian-etal-2018-polarity}, emotion intensity \cite{oladepo-etal-2025-predicting}, formality \cite{dementieva-etal-2023-detecting}, and concreteness \cite{Kewenig2025}. Scalar attributes also serve as controllable generation dimensions \cite{wang-demberg-2024-rsa}; we instead examine scoring artifacts under cross-domain evaluation.

\subsection{Domain Shift and Predictive Stability}

Distribution-shift theory relates target performance to source--target divergence \cite{ben2009theory}, and NLP models vary across domains in tagging \cite{peng-dredze-2017-multi} and classification \cite{zhou2024crossdomain}. Summarization benchmarks characterize variation across genre, style, and topic \cite{yuan-zhang-2025-domainsum,afzal-etal-2025-da}. Calibration can also change across contexts \cite{desai2020calibration}; similar cautions apply to scalar regression outputs. We complement that literature by separating contextual interpretation under fixed application and unsupervised target adaptation from usefulness in one specified decision.

\subsection{Technical and Domain-Specific Text}

Software documentation can carry, but also fragment, writer--reader understanding \cite{deboer2009documentation}; specificity has been associated with shared perspective in one human collaboration task \cite{hayashi2019specificity}; and human--agent research treats shared task representations as a basis for coordination \cite{scheutz2017shared}. Human--AI guidelines further support correction and user control \cite{amershi2019guidelines}. These works motivate the study of score transfer and an author-controlled candidate-ranking choice.

Technical and procedural documentation differs from general expository prose in lexical distribution and discourse organization. Procedural texts, in particular, frequently exhibit ordered steps, conditional branching, and goal-oriented structure \cite{aouladomar-saint-dizier-2005-towards}. Computational genre analyses likewise identify recurring schematic patterns that shape downstream modeling behavior \cite{chambers-jurafsky-2011-template}. Information retrieval research documents distinctive statistical properties of domain-specific corpora \cite{chen2018domain}. Software engineering research identifies structural variation and semantic gaps between documentation artifacts and related representations \cite{10.1145/3597503.3639130}. A 17-week production study paired runbook-grounded incident assistance with human-reviewed post-incident refinement and reported a downward trend in mean time to resolution, while noting that the uncontrolled setting precluded causal attribution \cite{dantonio2025hitl}. That study also used average TF--IDF as a rudimentary document-level specificity proxy, but found its small changes and term-level behavior difficult to interpret \cite{dantonio2025hitl}. A follow-on study reanalyzed weekly snapshots of the same runbook artifacts, rather than an independent deployment, using multiple linguistic and structural signals; their stability and direction varied across metrics and documentation contexts \cite{dantonio2026runbooks}. These findings motivate domain-aware examination of sentence-level specificity without treating any single signal as documentation quality.

\section{Data}

\subsection{Domain Selection}

Wikipedia explains; GitHub and Ansible often instruct; Python reference material defines. The study uses these visible differences to evaluate sentence-level specificity prediction across domains, contrasting general-domain exposition with technical and procedural documentation. General-domain text approximates the narrative and expository styles on which sentence specificity predictors were originally trained and evaluated, while the technical corpora foreground procedural instructions, parameter references, and definitional statements.

Corpora are treated as sources of naturally occurring sentences drawn from distinct stylistic domains.

\subsection{Corpora}

\textit{Wikipedia (General-Domain Reference).} English Wikipedia was extracted from publicly available dumps \cite{wikidump} using WikiExtractor \cite{wikiextractor}. Broad expository text written for a general audience makes this corpus the reference distribution for non-technical prose.

\textit{GitHub Platform Documentation.} The public \textit{github/docs} repository supplies primarily Markdown documentation for developer audiences \cite{githubdocs}. Its files combine conceptual explanations, API references, and procedural guidance.

\textit{Ansible Documentation.} From the public \texttt{ansible-documentation} repository \cite{ansibledocs}, we include configuration-oriented material, automation workflows, and parameter descriptions. Its sentences frequently depend on implicit execution context.

\textit{Python 3.12 Reference Documentation.} The official Python 3.12 documentation \cite{python312docs} contributes primarily definitional and reference-style text about language constructs and APIs. Lexical precision, rather than ordered task structure, distinguishes it from the procedural corpora.

\subsection{Preprocessing}

Corpora were processed independently. Executable code blocks were removed prior to sentence segmentation. Inline code spans and embedded technical identifiers were retained as lexical signals. Templating directives, structural fragments, and formatting artifacts were retained for the original large-scale distributional scoring condition.

Qualitative and controlled-edit sampling used deterministic annotatability filters; scored originals could retain markup, paths, and URLs. Sentence segmentation was performed using a uniform rule-based pipeline applied consistently across domains.

Retained artifacts offer one rival explanation for the corpus-level score differences. To test it, we select a \texttt{strict\_natural\_language\_v1} subset from the existing scored rows and apply the same surface rules to every corpus, without changing sentence boundaries, text, identifiers, or scores. The rules remove markup remnants, URLs, command- or path-dominated strings, marked heading or list fragments, strings shorter than five whitespace-delimited tokens, and strings with less than 60\% alphabetic content among non-space characters. This deliberately strict subset remains a sensitivity condition.

\subsection{Corpus Statistics}

All extracted sentences were included in large-scale scoring. Table~\ref{tab:corpus_stats} reports summary statistics after preprocessing.

\begin{table}[t]
\caption{Summary statistics for each corpus after preprocessing.}
\label{tab:corpus_stats}
\centering
\begin{tabular}{lcc}
\toprule
Domain & Sentences & Avg. Sent. Len \\
\midrule
Wikipedia & 900{,}406 & 21.96 \\
GitHub Docs & 170{,}798 & 15.24 \\
Ansible Docs & 44{,}452 & 11.73 \\
Python Ref & 179{,}549 & 11.64 \\
\bottomrule
\end{tabular}
\end{table}

Sentence counts are not size-matched. Descriptive statistics are computed at the sentence level, while bootstrap uncertainty resamples documents.

\section{Methodology}

\subsection{Questions and Tests}

\textit{Interpretation.} How do scores behave when their scoring artifacts are applied across general and technical domains? We compare distributions, corpus orders, and same-sentence ranks, then repeat after filtering and length standardization. Section~\ref{sec:reproducibility} records the exact implementations and repetitions.

\textit{Correspondence.} Within technical corpora, do the scores respond to controlled edits and agree locally with related granularity, human judgments, and matched LLM rubric judgments? These analyses test association and directional behavior rather than accuracy.

\textit{Decision value.} We test whether post-generation specificity scoring improves a bounded revision choice: does a score-based rule select a revision that matches the requested specificity change more often than the baseline of choosing the first generated candidate? For each source and requested change, a generator produces three alternatives without access to the scores, and the alternatives are fixed before scoring. One blinded evaluator compares each candidate with its source on a five-point scale; the midpoint denotes no perceived difference, and \texttt{X} marks an uninterpretable comparison that does not count as direction-valid. Each pre-specified rule selects one candidate, and paired resampling compares its success with the baseline on the same sources.

\subsection{Predictors and Comparator Identity}
\label{sec:predictors}

\textit{SpeciTeller.} The model is applied as a containerized, pinned dependency; Appendix Table~\ref{tab:component_provenance} records its component provenance. The released model is used without parameter modification or adaptation. Sentences are tokenized with NLTK's TreebankWordTokenizer before scoring to match the scorer's expected input format. Each sentence is scored independently, yielding a scalar specificity value where larger values indicate greater predicted specificity.

\textit{Pinned post-publication comparator.} Because the original 2019 checkpoint is unavailable, we use a later runnable implementation from the authors' official repository and treat it as a separate comparator rather than a reproduction of the published system. The implementation is pinned at the post-publication source revision listed in Appendix Table~\ref{tab:component_provenance}. The 2019 paper-era artifact provided neither a released checkpoint nor a matching official CPU path, so we selected the latest CPU-capable author-provided state and reproduced its legacy dependency stack in a pinned container without modifying the model.

Ko \textit{et al.}\ evaluated transfer from labeled news data to Twitter, Yelp reviews, and movie reviews \cite{ko2019domain}. A reproduction check on the released Twitter data and a supplemental three-run Yelp and Movie diagnostic used the pinned implementation's SE+AD mean--standard-deviation configuration. Across the three domains, observed Spearman values ranged from .262 to .614 against published values of .676--.750, Kendall's $\tau$ ranged from .184 to .444 against .487--.555, and MAE ranged from .139 to .164 against .107--.114. Rank correlations were lower and error was higher in all three domains. Appendix~\ref{sec:repro_details} reports exact values and provenance. A post-publication change also moved mean/standard-deviation distribution regularization from the labeled source batch to the unlabeled target batch, materially changing the objective.

The post-publication implementation is therefore reported as a separate comparator. Since the Python augmentation cannot be seeded, we repeat the same target adaptation three times per corpus to expose run variation, labeling the repetitions \textit{Adapt-1/2/3}. They are not different models, architectures, or conditions. The upstream seed controls NumPy and Torch, but not the Python augmentation. The secondary \textit{Adapt mean} averages all three scores per sentence. Both predictors score every sentence in the fixed corpus order.

\textit{GranuScore.} This measure represents semantic and referential granularity rather than descriptive specificity \cite{ellinger2026granuscore}. The official \texttt{granuscore==1.0.1} paper-default configuration assigns each sentence a 0--100 percentile, where higher is coarser or more abstract. Primary analyses retain this native direction. Because higher specificity scores instead indicate greater detail, agreement between the measures appears as a negative native correlation. For one secondary direction-aligned comparison, we therefore reverse only the sign of Spearman's $\rho$, so that positive values indicate agreement; the raw scores and ranks remain unchanged. The primary analysis retains sentences for which GranuScore extracts no referential unit and returns its sentinel value. A sensitivity analysis excludes those sentences to test whether they drive the result.

\subsection{Human-Reviewed Candidate Selection}
\label{sec:candidate_selection}

\begin{figure}[t]
\centering
\includegraphics[width=\columnwidth]{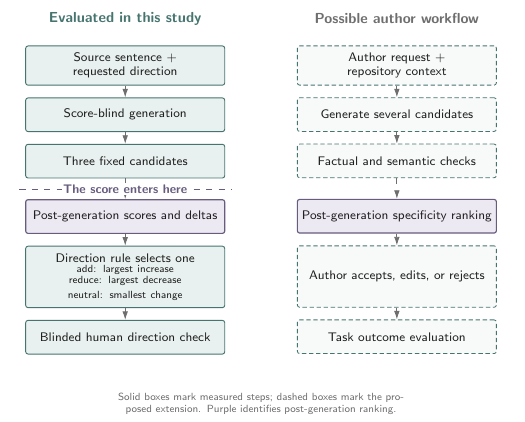}
\caption{Candidate selection as evaluated and as a possible documentation workflow. Left: three score-blind candidates are fixed before post-generation scoring, directional selection, and blinded review. Right: a proposed workflow could place the same ranking step after factual and semantic checks and before an author decision. Dashed boxes mark the proposed extension.}
\label{fig:operational_workflow}
\end{figure}

Candidate selection can be informed by post-generation ranking. For each source request (add, reduce, or neutral), a generator produces three alternatives without access to scores. Once fixed, a policy selects the largest score increase, largest decrease, or smallest absolute change, respectively. Blinded review tests whether this post-generation choice moved in the requested direction. Figure~\ref{fig:operational_workflow} shows the decision and one possible author-controlled use.

Pinned open-weight \texttt{gemma4:12b} and \texttt{gpt-oss:120b} models create two candidate sets from the same 60 technical sources (30 Ansible and 30 GitHub). Each produces three candidates per source, yielding 180 judgments per set, all rated by one evaluator without scores. Neither generator receives SpeciTeller, the pinned comparator, GranuScore, proxy values, prior candidates, or selection mappings. Actual admission uses only structured-response and text-integrity checks; every admitted candidate receives review. Separately, a counterfactual surface proxy checks direction-specific token-length change, retained content anchors, concrete markers, polarity, and modality after generation. It never excludes or alters candidates; it would reject 70 Gemma and 116 GPT-OSS-120B candidates. Appendix~\ref{sec:detailed_boundaries} reports its agreement with the blinded judgments.

One author evaluates each candidate set in a separate blinded session. Before review, we randomly assign the original and edited sentence to sides A and B and independently randomize the order of the comparison items. During rating, the evaluator is blind to source/candidate side, intended direction, all scores, proxy decisions, and selection policies. The scale runs from 1 (A clearly more specific) through 5 (B clearly more specific); 3 means no perceived difference, including a few limited-semantic-difference or alternative-interpretation cases, and X means uninterpretable. A selected candidate is direction-valid when the judgment matches the requested direction.

Only after all candidates are finalized do we compute SpeciTeller, each comparator repetition, and direction-aligned GranuScore. Each pre-specified policy selects by one score and is compared with always choosing the first candidate, the no-ranking baseline because candidate order is score-independent. Secondary policies use sentence-wise Adapt mean and, for GPT-OSS-120B, rank consensus across SpeciTeller, Adapt-1/2/3, and direction-aligned GranuScore. The consensus rule was specified before blinded human review, and its candidate selections were fixed before any ratings were collected. Ties select the earliest generated candidate. Human association and proxy analyses use all 180 ratings per set; each policy chooses once for each of 60 sources. To estimate gain uncertainty, 20{,}000 deterministic bootstrap samples resample sources within corpus and requested direction while keeping both policies paired. The review-order check reuses existing ratings. Candidate-set comparisons pair the same sources; separate review sessions make generator and session effects inseparable.

An illustrative case shows how the two policies operate without serving as representative evidence. For a neutral rewrite of ``When it is Inactive, this link will no longer be valid,'' the three score-blind candidates had SpeciTeller changes of $-.005299$, $+.004479$, and $+.000401$. The first-candidate baseline selected the first; the neutral SpeciTeller rule selected the third because its absolute change was smallest. Blinded review rated all three as unchanged in specificity, so both policies were direction-valid for this case. Appendix~\ref{sec:traceable_candidate_example} gives the case-ID prefix, candidate text, score changes, and blinded outcomes.

\subsection{Cross-Measure Comparison}

Do the score sources rank the same sentences and occupy their scales similarly? Within each corpus, Spearman's $\rho$ compares the rankings assigned to identical sentences. Raw values remain on each source's native scale: the predictors use 0--1 scores, GranuScore uses 0--100 coarseness percentiles, and the pilot uses 1--5 judgments; we never compare these raw values numerically across sources. Uncertainty uses 1{,}000 document-cluster bootstrap samples, keeping sentences from the same document together (seed 20260809). We report each comparator adaptation repetition's corpus mean separately. We also calculate the population SD across the three comparator scores for each sentence, then average those SDs across sentences. Strict filtering and exact-token procedures leave score definitions unchanged; for GranuScore, we repeat the analysis after excluding sentences with no extracted referential unit.

Distribution shape tests whether similar summaries conceal different scale use. We report native-scale quantiles and effective range $(p_{95}-p_{05})/(U-L)$ for bounds $[L,U]$. Boundary concentration is the share within 5\% of either theoretical limit; 1\% and 10\% thresholds test robustness. Empirical tails alone do not establish saturation. SpeciTeller and the three comparator repetitions remain primary, Adapt mean secondary, and the secondary GranuScore view excludes rows with no extracted referential unit.

\subsection{Preprocessing Sensitivity}

Could retained artifacts create the corpus pattern? We compare original rows with those retained by \texttt{strict\_natural\_language\_v1}. The predicate and thresholds are fixed before comparison and applied identically. For every excluded sentence, we record the applicable removal reason. We then use stable sentence identifiers to verify that retained and excluded counts exactly reconstruct the original row manifest. Selecting already scored rows neither alters text nor rescores inputs.

For each corpus, we report retention and original-versus-strict means and medians. We also compute changes in the Wikipedia-minus-technical mean and median gaps. Uncertainty is summarized with 1{,}000 nonparametric document-cluster bootstrap replicates within each corpus, using paired original/strict resampling, master seed 20260808, and 2.5th/97.5th percentile intervals. These intervals describe sensitivity to document composition.

\subsection{Length Standardization}

Are corpus scores different because sentence lengths differ? We analyze the original and strictly filtered data separately. Within each version, we retain exact token counts that are adequately represented in every corpus and reweight each corpus to the same token-count distribution. A count is eligible only when every corpus contributes at least 100 rows from at least 20 documents, yielding 1--38 tokens for original and 5--36 for strict rows. Over the eligible token counts, we normalize each corpus's length distribution and average the four distributions with equal corpus weight to form the reference distribution. We report weighted means, Wikipedia-minus-technical gaps, support retention, and Kish effective sample size. Uncertainty uses 1{,}000 within-corpus document-cluster bootstraps with a fixed reference, master seed 20260809, and percentile intervals.

As a sensitivity check, we fit ordinary least squares models with corpus indicators and a five-knot restricted cubic spline of log token count on the same support \cite{harrell2015regression}, using document-cluster-robust covariance \cite{cameron2015cluster}. To test whether an adjusted corpus gap changes with sentence length, we evaluate the pre-specified corpus-by-restricted-spline interaction at five quantiles of the reference length distribution. We classify the gap as meaningfully nonconstant if these estimates span at least 0.10 score units or change sign. Character count is not jointly controlled. These analyses describe residual cross-corpus model-score differences after observed token-length adjustment.

\subsection{Proxy Feature Alignment}

Which visible cues travel with a score in each domain? We treat the cues as transparent audit indicators, not as alternative measures of specificity. Sentence length and lexical rarity are motivated by the shallow feature families used by SpeciTeller and the comparator, which include length and summary IDF statistics \cite{li2015speciteller,ko2019domain}; IDF relates term informativeness to corpus rarity \cite{SparckJones1972}. Terms that occur in relatively few sentences may identify particular entities or concepts and thus provide one observable cue to detail, although rarity alone is not specificity. We operationalize lexical rarity within each corpus as the mean and maximum TF--IDF across the unique tokens in each sentence. Token count operationalizes length, while the proportion of technical or code-like tokens provides a documentation-oriented surface cue. Within each corpus, Spearman correlations quantify the rank association between each surface cue and the specificity score. Deterministic equal-count bins display how scores vary across the cue's range. A secondary decomposition separates identifiers, paths or flags, version and numeric tokens, parameter assignments, and token-shape complexity.

\subsection{Controlled Edit Behavioral Probing}

Local sensitivity under cross-domain application is examined through a controlled experiment with author-crafted edits on two technical corpora. Sentences are deterministically sampled from the lowest specificity decile, a median band, and the highest decile, with token-length constraints of 8--40 tokens. Each sentence is manually rewritten under one of three edit types: increasing descriptive specificity, reducing descriptive specificity, or paraphrasing without systematic change. Edits function as behavioral probes.

For each edited sentence, $\Delta = \text{score}_{edited} - \text{score}_{original}$ is computed. We report mean $\Delta$, median $\Delta$, mean absolute $\Delta$, and interquartile range. Directional consistency is defined as $\Delta > 0$ for specificity-increasing edits, $\Delta < 0$ for specificity-reducing edits, and $|\Delta| \leq 0.05$ for neutral paraphrases.

The same 60 author-crafted pairs receive GranuScore without new selection: expected native deltas are negative for adding specificity and positive for de-specifying. Paired intervals use 10{,}000 within-cell bootstraps.

\subsection{Human Pilot Annotation}

For each of two technical corpora, a deterministic 40-sentence sample is drawn from SpeciTeller-score extremes and a median band. Each row contains the sentence text, a blank label field, an optional note field, and a stable sentence identifier; model scores are not shown to annotators. Three adult annotators independently assign ordinal specificity labels on a five-point scale from very vague or very general to very specific or highly concrete. The annotation instructions define specificity in terms of concrete details, identifiable entities, and explicit operational information; the instructions are summarized in Appendix~\ref{sec:pilot_binned}.

Human labels anchor the final correspondence test. We report pairwise agreement and quadratic-weighted Cohen's $\kappa$ \cite{cohen1968weighted}, then correlate individual and pooled labels with fixed SpeciTeller, the three comparator repetitions, and GranuScore by stable sentence identifier. Adapt mean is secondary and non-independent. GranuScore enters the pilot table only through a secondary direction-aligned rank view; its native coefficient has equal magnitude and opposite sign.

Matched LLM judges test whether the pattern extends beyond one rubric judge. Qwen3-14B \cite{yang2025qwen3}, \texttt{gemma4:12b}, \texttt{gpt-oss:20b}, and \texttt{gpt-oss:120b} apply the same rubric to the same 80 pilot sentences, one blind sentence per fresh conversation, with deterministic decoding and a strict integer schema. Qwen and the remote GPT-OSS-120B run use Ollama 0.32.5. Local replications use Ollama 0.32.9 to meet model-interface requirements; all four judges perform the same task. Each local judge scores three non-project test sentences three times to check repeated-run agreement. The 120B rubric scores are evaluated only against the pilot labels, not its candidates.

Score--human correlations, pairwise score-source agreement, and score--human correlation contrasts are computed separately on the selected 40 GitHub and 40 Ansible rows. Uncertainty uses 10{,}000 sentence-bootstrap replicates, jointly resampling all scores and labels within corpus with master seed 20260809. The non-expert ordinal labels provide descriptive correspondence.

\section{Results}

We report decision value first (A), followed by interpretation (B--F) and correspondence (G--H).

\subsection{Decision-Level Candidate Selection}

Table~\ref{tab:candidate_selection} compares both candidate sets using exploratory, unadjusted intervals. Among the individual score rules, only SpeciTeller on the Gemma set has a gain interval excluding zero; the secondary consensus does so on the GPT-OSS-120B set. All primary GPT-OSS-120B and between-set gain intervals include zero.

\begin{table*}[t]
\caption{Post-generation ranking by generator and review session. Each cell reports direction-valid selections and the paired percentage-point gain over that generator's first candidate, with 95\% source-case bootstrap intervals. Adapt-1/2/3 repeat one pinned implementation; Adapt mean and consensus are secondary.}
\label{tab:candidate_selection}
\centering
\scriptsize
\setlength{\tabcolsep}{4.2pt}
\begin{tabular}{lcccc}
\toprule
& \multicolumn{2}{c}{Gemma} & \multicolumn{2}{c}{GPT-OSS-120B} \\
\cmidrule(lr){2-3}\cmidrule(lr){4-5}
Policy & Direction-valid & Gain [95\% CI] & Direction-valid & Gain [95\% CI] \\
\midrule
First candidate & 43/60 (71.7\%) & reference & 31/60 (51.7\%) & reference \\
SpeciTeller & 50/60 (83.3\%) & +11.7 [+1.7, +21.7] & 34/60 (56.7\%) & +5.0 [$-6.7$, +16.7] \\
Adapt-1 & 46/60 (76.7\%) & +5.0 [$-3.3$, +13.3] & 34/60 (56.7\%) & +5.0 [$-6.7$, +16.7] \\
Adapt-2 & 45/60 (75.0\%) & +3.3 [$-6.7$, +13.3] & 34/60 (56.7\%) & +5.0 [$-6.7$, +16.7] \\
Adapt-3 & 45/60 (75.0\%) & +3.3 [$-6.7$, +13.3] & 36/60 (60.0\%) & +8.3 [$-1.7$, +20.0] \\
Adapt mean (sec.) & 45/60 (75.0\%) & +3.3 [$-6.7$, +13.3] & 35/60 (58.3\%) & +6.7 [$-5.0$, +18.3] \\
Aligned GS & 48/60 (80.0\%) & +8.3 [0.0, +16.7] & 33/60 (55.0\%) & +3.3 [$-6.7$, +13.3] \\
Consensus (sec.) & n/a & n/a & 38/60 (63.3\%) & +11.7 [+1.7, +21.7] \\
\bottomrule
\end{tabular}
\end{table*}

The Gemma SpeciTeller gain concentrates in add-specific requests, while the review-order check retains its direction with less precision. The GPT-OSS-120B set has lower overall direction-validity and higher proxy false-rejection; despite positive score--human association, its individual-score selection gains remain uncertain. Appendix~\ref{sec:detailed_boundaries} gives the request-specific, review-order, association, and proxy diagnostics.

\subsection{Predictor-Dependent Corpus Patterns}

Table~\ref{tab:cross_context_profile} shows that corpus order and same-sentence ranks depend on score source: the two specificity sources reverse the Wikipedia--GitHub and Wikipedia--Ansible contrasts, while technical-corpus rank agreement ranges from negative to moderate. Appendix Table~\ref{tab:predictor_details} reports repetition means, uncertainty intervals, and native-scale gaps.

\begin{table*}[t]
\caption{Cross-context measurement profile. Higher ST/Adapt means more specific; higher native GS means coarser. Means retain native scales; other entries are within-corpus Spearman correlations. Adapt mean is secondary; row SD summarizes variation across three repetitions.}
\label{tab:cross_context_profile}
\centering
\scriptsize
\setlength{\tabcolsep}{6.0pt}
\begin{tabular}{lrrrr}
\toprule
Metric & Wikipedia & GitHub Docs & Ansible Docs & Python Ref \\
\midrule
\multicolumn{5}{l}{\textit{Score level, variability, and agreement}} \\
SpeciTeller mean & .7534 & .4960 & .4257 & .2921 \\
Adapt mean & .3490 & .3993 & .3704 & .2825 \\
Adapt row SD & .0239 & .0324 & .0171 & .1542 \\
ST--Adapt mean $\rho$ & .5103 & .0704 & $-.0656$ & .4430 \\
\midrule
\multicolumn{5}{l}{\textit{Related granularity measure}} \\
GranuScore mean & 57.07 & 68.87 & 68.19 & 67.50 \\
GS--ST native $\rho$ & $-.137$ & $-.282$ & $-.281$ & $-.364$ \\
GS--Adapt mean native $\rho$ & .237 & .151 & .045 & $-.202$ \\
\midrule
\multicolumn{5}{l}{\textit{SpeciTeller association with visible cues}} \\
Mean TF--IDF $\rho$ & $-.729$ & $-.387$ & $-.229$ & $-.167$ \\
Technical-token ratio $\rho$ & .359 & .424 & .374 & .334 \\
Token count $\rho$ & .754 & .417 & .312 & .242 \\
\bottomrule
\end{tabular}
\end{table*}

Python stands out for comparator repetition variability. Separate range and boundary diagnostics likewise show source-specific scale use rather than uniform saturation (Appendix~\ref{sec:distribution_details}).

\subsection{Related Granularity Measure}

GranuScore retains a distinct corpus order and correlates negatively with SpeciTeller in every corpus; its comparator relationship changes sign in Python. Filtering, unit-only sensitivity, and exact-token standardization preserve its corpus ordering (Appendix~\ref{sec:granuscore_details}).

\subsection{Preprocessing Sensitivity}

Strict filtering preserves Wikipedia as highest and Python as lowest and widens every Wikipedia-to-technical mean and median gap. Table~\ref{tab:preprocessing_sensitivity} compares the original summaries with this deliberately strict row-selection condition. Retention is low and uneven: 35.6\% for Wikipedia, 46.1\% for GitHub, 48.5\% for Ansible, and 70.9\% for Python. GitHub and Ansible exchange their middle positions, so the complete four-corpus ordering does not remain unchanged.

\begin{table}[t]
\caption{Preprocessing sensitivity on original and strict subsets. ``Keep'' is the percentage retained by the strict condition.}
\label{tab:preprocessing_sensitivity}
\centering
\setlength{\tabcolsep}{2.5pt}
\scriptsize
\begin{tabular}{lrrrrr}
\toprule
Corpus & Keep & \multicolumn{2}{c}{Mean} & \multicolumn{2}{c}{Median} \\
\cmidrule(lr){3-4}\cmidrule(lr){5-6}
 & (\%) & Orig. & Strict & Orig. & Strict \\
\midrule
Wikipedia & 35.6 & 0.753 & 0.764 & 0.884 & 0.942 \\
GitHub Docs & 46.1 & 0.496 & 0.340 & 0.543 & 0.234 \\
Ansible Docs & 48.5 & 0.426 & 0.388 & 0.369 & 0.298 \\
Python Ref & 70.9 & 0.292 & 0.284 & 0.154 & 0.127 \\
\bottomrule
\end{tabular}
\end{table}

Mean-gap changes are $+0.167$ for GitHub (95\% document-cluster bootstrap interval $[0.160, 0.175]$), $+0.049$ for Ansible ($[0.037, 0.060]$), and $+0.020$ for Python ($[0.013, 0.026]$). Median-gap changes are $+0.367$ ($[0.356, 0.378]$), $+0.129$ ($[0.110, 0.151]$), and $+0.085$ ($[0.073, 0.096]$), respectively. Thus, all six gap-change intervals lie above zero under this sensitivity condition.

The comparator is also sensitive to the same score-blind selection, but it does not converge toward SpeciTeller. Its strict-row Adapt-mean order is Ansible, Wikipedia, GitHub, then Python; the strict Wikipedia-minus-technical gaps are $+0.0089$ for GitHub, $-0.0341$ for Ansible, and $+0.1983$ for Python. Filtering therefore changes each predictor's corpus pattern without reconciling them.

\subsection{Length-Controlled Comparisons}

Exact-token standardization reduces every Wikipedia-to-technical SpeciTeller gap in both the original and strictly filtered data, but all remain positive. Table~\ref{tab:length_control} compares the raw gaps with gaps standardized to a shared token-count distribution. All document-cluster bootstrap intervals exclude zero, and adjustment retains 49.1\%--94.5\% of each raw magnitude. Under this design, observed token length therefore explains part, but not all, of the corpus-level score differences.

\begin{table}[t]
\caption{Raw and token-length-standardized Wikipedia-minus-technical SpeciTeller gaps. Brackets give 95\% document-cluster bootstrap intervals; ``Ret.'' is the percentage of raw gap magnitude retained.}
\label{tab:length_control}
\centering
\setlength{\tabcolsep}{2.2pt}
\scriptsize
\begin{tabular}{llrrr}
\toprule
Rows & Corpus & Raw & Standardized [95\% CI] & Ret. \\
\midrule
Orig. & GitHub & 0.257 & 0.243 [0.236, 0.249] & 94.5\% \\
Orig. & Ansible & 0.328 & 0.271 [0.253, 0.290] & 82.6\% \\
Orig. & Python & 0.461 & 0.329 [0.303, 0.355] & 71.3\% \\
Strict & GitHub & 0.425 & 0.304 [0.295, 0.313] & 71.6\% \\
Strict & Ansible & 0.376 & 0.258 [0.236, 0.281] & 68.5\% \\
Strict & Python & 0.481 & 0.236 [0.208, 0.264] & 49.1\% \\
\bottomrule
\end{tabular}
\end{table}

Overlap qualifies the attenuation. Common-support retention is 83.0\% for Wikipedia in the original rows and 77.0\% in the strict rows. Python has Kish effective-sample-size fractions of 0.559 and 0.622, with maximum weights of 8.241 and 6.493, respectively. These diagnostics qualify the adjusted estimates even though the pre-specified weight threshold is met.

Length explains part of the gaps, not one constant share. The interaction diagnostic is substantial for all six contrasts. Across the five evaluated counts (2, 5, 10, 17, and 26 tokens), original-row point gaps range from 0.067--0.353 for GitHub, 0.123--0.420 for Ansible, and 0.112--0.504 for Python. Across 7, 10, 14, 21, and 28 tokens in the strict rows, the corresponding ranges are 0.222--0.418, 0.129--0.396, and 0.043--0.411. Every evaluated point gap remains positive, but the material variation supports heterogeneous persistence rather than one invariant adjusted effect. The additive spline sensitivity likewise yields positive average gaps, while interaction partial $R^2$ values of 0.050 and 0.033 reinforce the length-specific interpretation.

Applying the identical standardization to Adapt mean yields original-row gaps of $-0.0934$, $-0.1103$, and $-0.0085$ for GitHub, Ansible, and Python. Strict-row gaps are $-0.0611$, $-0.0990$, and $+0.1053$. All original comparator interaction diagnostics are substantial, and several contrasts change sign or magnitude after adjustment. Exact-token standardization thus changes both predictors' patterns but does not make their orderings or contrasts agree.

\subsection{Feature Alignment Across Domains}

The visible-cue rows in Table~\ref{tab:cross_context_profile} show strong length and TF--IDF associations in Wikipedia but weaker associations in technical corpora, where technical-token ratio remains positive. Appendix~\ref{sec:appendix_robustness_proxy} reports the full feature table and shows that the aggregate technical-token association is not one uniform code-like signal.

\subsection{Controlled Edit Behavioral Probing}

Across 60 author-crafted pairs, both measures respond locally. SpeciTeller moves as intended for all GitHub add- and de-specific edits and most Ansible edits; one Ansible edit adds a duplicated ``such as'' clause and ``02d 23h 58m 59s'', raising its score from .277 to .595 ($\Delta=+.318$). GranuScore moves in its expected native direction, while both irrelevant-rewrite intervals include zero. Appendix Table~\ref{tab:controlled_edits} gives full results.

\subsection{Human Pilot Annotation}

This selected 40-sentence-per-corpus pilot is descriptive. Agreement among the three non-expert annotators is low to moderate and corpus-dependent: quadratic-weighted Cohen's $\kappa$ is approximately .20--.30 in Ansible and .07--.49 in GitHub. Against the pooled labels, SpeciTeller has positive rank correspondence in both corpora ($\rho=.480$ [.126, .723] for Ansible and $.715$ [.518, .844] for GitHub), but these estimates characterize local association under this sample and rubric, not accuracy or general human validity. Correspondence across the other score sources varies by corpus, annotator, source, and judge. All Qwen-versus-judge intervals in GitHub include zero, while several Ansible contrasts do not. Appendix Table~\ref{tab:human_alignment} reports all coefficients and intervals; Appendix~\ref{sec:pilot_binned} gives the score-source contrasts and annotator-specific ranges.

\section{Discussion}

Our results follow the three questions introduced in the Introduction. Interpretation asks what a sentence-specificity score means under cross-domain evaluation. Correspondence asks whether it responds to targeted changes and agrees with related measures and readers. Decision value asks whether the score improves one bounded post-generation choice. Following these stages in order distinguishes score magnitude from cross-domain interpretation, correspondence from decision value, and ranking from candidate admission.

\subsection{Interpretation}

The interpretation analyses show that sentence-specificity scores are predictor- and domain-dependent. Across one general-domain and three technical corpora, predictor choice changes both corpus ordering and the ranking of identical sentences. The combined evidence in Tables~\ref{tab:cross_context_profile}--\ref{tab:length_control} shows that filtering and exact-token adjustment change magnitudes but do not reconcile those differences. Scale use and adaptation variability also remain source-dependent, while visible associations shift from general expository cues toward heterogeneous technical-token cues. The divergences therefore cannot be reduced to one preprocessing choice or to sentence length alone.

These checks narrow the explanation of the observed divergences. Filtering changes their magnitude, and token length accounts for part of the variation, while corpus ordering, rank agreement, scale behavior, and cue associations continue to vary by predictor and corpus. Corpus means therefore summarize a score source's behavior in a particular corpus, and threshold-based uses inherit that predictor- and corpus-specific interpretation.

\subsection{Correspondence}

The second stage moves from global interpretation to local behavior. Controlled edits test directional response, GranuScore tests overlap with related semantic and referential granularity, and human and matched LLM judgments test correspondence with readers applying the same rubric. Both SpeciTeller and GranuScore respond in their expected directions to the author-crafted edits, but the SpeciTeller response is smaller and less consistent for Ansible additions than for GitHub additions. GranuScore remains related but non-interchangeable: its correspondence changes across corpus and score source rather than supplying a second calibration of specificity.

Within the selected 40-sentence-per-corpus pilot, the three non-expert annotators show low-to-moderate, corpus-dependent agreement ($\kappa\approx.20$--$.30$ in Ansible and $.07$--$.49$ in GitHub). Against their pooled labels, SpeciTeller correlates positively in both corpora (Appendix Table~\ref{tab:human_alignment}). Under the same rubric, Spark-hosted GPT-OSS-120B shows no consistent correspondence advantage over the 12B--20B judges. Separately, its generated candidate set has lower direction-validity than Gemma's. The selected sample, variable agreement, judge contrasts, and inseparable candidate-set/session effects bound these results; the stage establishes local correspondence, not comparative accuracy, expert validation, or decision value.

\subsection{Decision Value}

The decision analysis tests whether an ordering identifies a direction-valid choice from fixed alternatives. The Gemma SpeciTeller gain is +11.7 points [95\% CI +1.7, +21.7]; the GPT-OSS-120B estimate is +5.0 points [$-6.7$, +16.7] (Table~\ref{tab:candidate_selection}). We did not specify a minimum practically important gain before analysis. Every primary single-score GPT-OSS-120B interval includes zero despite positive score--human association; only the secondary consensus interval excludes zero. Positive correspondence therefore does not guarantee selection improvement, and the Gemma result supports a bounded ranking decision rather than a predictor-independent guarantee.

The concentration of the SpeciTeller gain in add-specific requests shows that the observed benefit is direction- and candidate-set-specific; it does not establish that one direction is generally easier. The high false-rejection rate of the counterfactual surface proxy also shows that cues useful for explaining a score need not be safe candidate filters. Local correspondence or edit sensitivity cannot replace this decision-level test.

\subsection{Three-Stage Evidence and Context-Aware Reuse}

Together, these analyses show how the three-stage evidence structure can be reused to ask, in turn, what a score represents in the target context, whether it behaves and corresponds as expected, and whether it improves the named decision.

The evaluated decision is limited to post-generation ordering of fixed candidates for direction validity. In a possible collaborative-documentation workflow, factual and semantic checks could precede that ranking and a person could retain the final task-level decision. Figure~\ref{fig:operational_workflow} presents those checks and the author decision as an unevaluated extension. We do not test author adoption or heuristics such as choosing the longest candidate; the baseline is the score-independent first candidate. Scores are consequently most defensible for comparisons within a characterized document family and decision, rather than for cross-corpus thresholds or automatic rejection.

Factual inclusion, meaning preservation, and task completion remain separate decisions from the evaluated sentence-specificity ranking. The study does not establish whether locally valid score movements improve the context available to people or AI systems.

The cross-domain results also motivate a deeper measurement question: whether technical-documentation specificity needs a better scalar model or a more explicit multidimensional account of procedures, identifiers, constraints, and entities. On the selected pilot, Qwen's pooled correlation with non-expert labels is $\rho=.653$ [.356, .853] in Ansible and $.611$ [.352, .807] in GitHub, while direction-aligned GranuScore yields $.612$ [.340, .796] and $.017$ [$-.259$, .300], respectively (Appendix Table~\ref{tab:human_alignment}). These preliminary, corpus-dependent correlations make LLM rubrics and related granularity measures concrete research directions. Comparing them under the same three-stage evidence structure would preserve the distinction among interpretation, correspondence, and decision value.

\section{Conclusion}

Our findings give a bounded answer about applying SpeciTeller from general prose to technical documentation. SpeciTeller assigns lower mean scores to all three technical corpora than to Wikipedia. Strict filtering changes the gaps and exact-token standardization reduces them, but every standardized Wikipedia-to-technical SpeciTeller gap remains positive. The pinned post-publication author-repository implementation used as the comparator instead reverses the Wikipedia--GitHub and Wikipedia--Ansible contrasts on the original rows, while same-sentence agreement between the predictors is weak in both corpora. The resulting corpus order is therefore predictor- and document-family-specific rather than a shared cross-domain scale.

Within the evaluated technical settings, SpeciTeller nevertheless provides a locally informative signal: its scores correlate positively with pooled non-expert judgments and respond in the intended direction to controlled edits. Its candidate-selection gain is +11.7 points [+1.7, +21.7] for Gemma and +5.0 points [$-6.7$, +16.7] for GPT-OSS-120B; every primary single-score GPT-OSS-120B interval includes zero. The evaluated pinned target-adapted implementation changes the measured ordering without producing a uniformly stronger correspondence or selection rule. Scores therefore need characterization within the document family and testing on the decision they will inform.

Together, these results separate score interpretation, behavioral and judgment correspondence, and value for a named decision. Across the predictors and related measures evaluated here, performance remains corpus- and decision-dependent, leaving room to improve specificity measurement for technical documentation. Future work will evaluate documentation-specific training, calibration, and multidimensional representations under larger expert review and executable tasks.

\section{Reproducibility}
\label{sec:reproducibility}

The SpecTech source repository is available at \url{https://github.com/DocSpectra/SpecTech}. The workflow includes commands to reconstruct the corpora and rerun scoring, generation, and analysis. Its README provides stage-level entry commands, while fixed configurations, row manifests, and data schemas document each run's source snapshot and sentence order, software, model, and runtime versions, exact prompts and schema versions, random seeds, join definitions, and required checksums. Validation checks reject mismatches in source or component hashes, row identities, or row order; incomplete coverage; invalid schemas; and joins that do not account for every row. Appendix Tables~\ref{tab:corpus_provenance}--\ref{tab:evidence_provenance} provide typed verification identifiers, and Appendix~\ref{sec:llm_prompt_roles} reproduces the generation and rubric prompts.

GPT-OSS-120B candidate generation and rubric judging ran on an NVIDIA DGX Spark with a GB10 Grace Blackwell Superchip and 128~GB of unified memory. All other experimental work ran on an Intel Core i9-14900KF workstation with an NVIDIA GeForce RTX 4080 SUPER.

\section*{Limitations}

Without sentence-level gold labels for the project corpora, the predictors' native scores remain uncalibrated to a shared scale. Score gaps and agreement therefore characterize relative behavior, not comparative accuracy.

Candidate selection is bounded to one evaluator, two separate review sessions, the same 60 source cases with three candidates each, two corpora, two generators, and one pre-specified direction-validity rule. The observed arm difference combines candidate-set and session effects.

Comparator identity carries a separate limit: it is a pinned post-publication implementation of Ko \textit{et al.}'s predictor, not a reproduced 2019 checkpoint or an author-designated replacement for the 2019 Table~2 system. Released-data checks on Twitter, Yelp, and Movie all produced lower rank correlations and higher MAE than the published results, but the unavailable paper-era artifact prevents attribution of the reproduction gap to a particular implementation difference. The changed objective, unseeded augmentation, and unresolved Python run variation limit the project findings to this target-adapted implementation and its Adapt-1/2/3 repetitions.

As a related construct, GranuScore measures semantic and referential granularity rather than descriptive specificity. Some rows receive its exact no-unit sentinel, which the primary analysis retains and a sensitivity analysis excludes. GranuScore therefore provides a construct-sensitive comparison, not a third specificity predictor.

Some sampled originals and author-crafted edits retain source-format remnants or typographical slips; all were scored as written, so deltas may reflect formatting and token changes as well as descriptive detail.

Strict filtering keeps only 35.6\%--70.9\% of rows and changes the four corpora by unequal amounts. Wider or reordered gaps after filtering can reflect the changed samples, so this condition is a sensitivity check rather than a cleaner replacement dataset.

Length adjustment omits character count and other differences; Python has the weakest common support despite meeting the weight threshold. Adjusted gaps retain other confounding and do not establish a causal or invariant domain effect.

Visible lexical and token-shape cues form the proxy features; they are not semantic, syntactic, ontological, or model-internal explanations. Their correlations can show what travels with a score, but they cannot tell us why the model produced it or what the sentence means.

The 40+40 pilot uses selected sentences, three non-expert annotators, and one score per judge and sentence.

Coverage stops at one general English corpus and three English technical-documentation corpora, scored by SpeciTeller and one pinned post-publication target-adapted comparator. Other documentation types, languages, and predictor families may show different score behavior.

Evaluation is limited to existing language scores and one candidate-selection decision across two candidate sets. Figure~\ref{fig:operational_workflow} shows a possible author-controlled use, but its contextual checks, author choice, and task outcomes remain unevaluated. Retrieval, documentation quality, user, team, and collaboration outcomes remain outside the claim scope.

\section*{Ethical Considerations}

This study evaluates existing linguistic scoring methods across public technical-documentation corpora and English Wikipedia. Three adult annotators were recruited through personal networks and participated voluntarily. They were informed that their anonymized responses would be used for research, provided consent before participating, and received nominal nonmonetary compensation. The procedure posed minimal risk, and no personal or sensitive information was collected. Under institutional guidelines, the study qualified as exempt from formal IRB review.

\section*{AI Disclosure}

OpenAI Codex assisted with manuscript drafting and editing, programming, and development of portions of the statistical analysis. The authors reviewed and approved the resulting research design, verified the implementation and reported results, and assume responsibility for the work.

\clearpage
\bibliographystyle{IEEEtran}
\bibliography{custom}

\clearpage
\appendices
\section{Illustrative Controlled Edit Examples}
\label{sec:appendix_edits}

\newcommand{\storedbacktick}{\pdfliteral direct{/Span <</ActualText (`)>> BDC}\texttt{\char0}\pdfliteral direct{EMC}}

The examples below show illustrative author-crafted edits used in the behavioral probing analysis.

\begin{table}[H]
\caption{Controlled edit score deltas, $\Delta$, and directional consistency.}
\label{tab:controlled_edits}
\centering
\setlength{\tabcolsep}{3.5pt}
\small
\begin{tabular}{llccc}
\toprule
Corpus & Edit & Mean $\Delta$ & Med. $\Delta$ & Cons. \\
\midrule
GitHub & add\_spec & 0.124 & 0.036 & 100\% \\
GitHub & de\_spec & -0.180 & -0.162 & 100\% \\
GitHub & irrelev. & -0.001 & -0.003 & 80\% \\
\midrule
Ansible & add\_spec & 0.054 & 0.020 & 75\% \\
Ansible & de\_spec & -0.161 & -0.023 & 91.7\% \\
Ansible & irrelev. & -0.103 & -0.005 & 80\% \\
\bottomrule
\end{tabular}
\end{table}

GranuScore moves in its expected native direction. Adding specificity lowers its mean by 6.43 [$-12.76$, $-0.54$] in Ansible and 2.52 [$-3.54$, $-1.55$] in GitHub; de-specifying raises it by 3.64 [1.19, 6.29] and 5.38 [1.16, 10.50]. Both irrelevant-rewrite intervals include zero.

\subsection{Add-Specific Example from Ansible}

\textit{Original} (0.277): \detokenize{An example to show a host uptime in days/hours/minutes/seconds (assuming facts were gathered).}

\textit{Edited} (0.595): \detokenize{An example to show a host uptime in days/hours/minutes/seconds (assuming facts were gathered) such as such as}\,\textsuperscript{\textit{[sic]}}\,\detokenize{02d 23h 58m 59s.}

$\Delta = +0.318$

\subsection{De-Specify Example from Ansible}

\begingroup
\raggedright
\textit{Original} (0.960): * They SHOULD be listed in corresponding files under the \storedbacktick\storedbacktick meta\storedbacktick\storedbacktick{} directory, specified in \storedbacktick\storedbacktick\nolinkurl{meta/execution-environment.yml}\storedbacktick\storedbacktick, and \storedbacktick verified <\nolinkurl{https://ansible.readthedocs.io/projects/builder/en/latest/collection_metadata/#when-installing-collections-using-ansible-galaxy}>\storedbacktick\_. \textit{(list marker retained from the source line)}
\par
\endgroup

\textit{Edited} (0.127): \detokenize{They SHOULD be listed in corresponding files under the meta directory and verified.}

$\Delta = -0.833$

\subsection{Irrelevant Rewrite Example from Ansible}

\textit{Original} (0.016): In other words, the different variables will override each other in a certain order.

\textit{Edited} (0.013): In a certain order, different variables will override each other.

$\Delta = -0.003$

\section{Pilot Score--Label Binned Correspondence}
\label{sec:pilot_binned}

\subsection{Pilot Annotation Instructions}

Annotators were asked to add a number in the \texttt{human\_label} column for each sentence using a five-point scale: 1 = very vague or very general, 2 = somewhat vague, 3 = moderately specific, 4 = specific, and 5 = very specific or highly concrete. The instructions described a sentence as more specific when it mentions concrete details such as versions, file paths, commands, parameters, identifiers, numbers, identifiable entities, or explicit operational information. They described a sentence as less specific when it uses general language, makes high-level statements without details, or lacks concrete identifiers. The instructions asked annotators to use their first impression and noted that there were no right or wrong answers. As an illustrative contrast, the instructions presented ``Restart the service'' as likely lower specificity and ``Restart the nginx service using systemctl restart nginx on host web-03'' as likely higher specificity. The task was estimated to take about 10--15 minutes total.

\subsection{Unified Pilot Correlations}

\begin{table*}[t]
\caption{Unified pilot Spearman correlation ($n=40$ per corpus). ST and comparator Adapt-1/2/3 are specificity predictors; these are repeated fits of the same pinned post-publication implementation. Qwen, Gemma, GPT-OSS-20B, and GPT-OSS-120B are matched LLM rubric judges. Pooled cells give paired-bootstrap 95\% intervals. Adapt mean and GS-align are secondary; GS-align negates native GranuScore rank association and is not a raw-score conversion.}
\label{tab:human_alignment}
\centering
\setlength{\tabcolsep}{5pt}
\scriptsize
\begin{tabular}{llrrrr}
\toprule
Corpus & Score instance & Ann. A & Ann. B & Ann. C & Pooled $\rho$ [95\% CI] \\
\midrule
Ansible & ST    & .188 & .502 & .733 & .480 [.126, .723] \\
        & Adapt-1 & .302 & .254 & .604 & .484 [.189, .686] \\
        & Adapt-2 & .304 & .256 & .600 & .486 [.194, .688] \\
        & Adapt-3 & .305 & .253 & .615 & .486 [.183, .696] \\
        & Adapt mean & .304 & .249 & .600 & .480 [.184, .686] \\
        & Qwen  & .408 & .578 & .656 & .653 [.356, .853] \\
        & Gemma & .159 & .526 & .717 & .481 [.144, .723] \\
        & GPT-OSS-20B & .021 & .436 & .786 & .352 [$-.018$, .625] \\
        & GPT-OSS-120B & .077 & .568 & .772 & .455 [.110, .689] \\
        & GS-align & .409 & .476 & .572 & .612 [.340, .796] \\
\midrule
GitHub  & ST    & .312 & .506 & .765 & .715 [.518, .844] \\
        & Adapt-1 & .392 & .308 & .152 & .436 [.173, .650] \\
        & Adapt-2 & .501 & .233 & .220 & .497 [.245, .688] \\
        & Adapt-3 & .360 & .326 & .168 & .415 [.147, .629] \\
        & Adapt mean & .399 & .302 & .173 & .448 [.189, .654] \\
        & Qwen  & .098 & .577 & .773 & .611 [.352, .807] \\
        & Gemma & .119 & .565 & .822 & .673 [.449, .819] \\
        & GPT-OSS-20B & .034 & .606 & .707 & .589 [.351, .762] \\
        & GPT-OSS-120B & .028 & .596 & .787 & .594 [.345, .780] \\
        & GS-align & $-.169$ & .054 & .245 & .017 [$-.259$, .300] \\
\bottomrule
\end{tabular}
\end{table*}

The ST-minus-comparator pooled contrasts are $-.004$ [$-.267$, .220], $-.007$ [$-.277$, .221], and $-.007$ [$-.271$, .216] for Adapt-1/2/3 in Ansible, and $+.279$ [.030, .540], $+.218$ [$-.012$, .455], and $+.300$ [.065, .549] in GitHub.

All direction-aligned GranuScore-minus-comparator intervals include zero in Ansible. In GitHub, aligned GranuScore differs from SpeciTeller, Adapt-2, and Qwen, but not Adapt-1 or Adapt-3. Qwen-minus-specificity-source differences range from $+0.166$ to $+0.173$ in Ansible and $-0.104$ to $+0.196$ in GitHub; all eight intervals include zero.

On these selected rows, Qwen--SpeciTeller rank agreement is $0.771$ in Ansible and $0.533$ in GitHub, while Qwen--comparator agreement is $0.289$--$0.315$ and $0.265$--$0.296$. Adapt-1/2/3 pairwise $\rho$ remains $0.983$--$0.997$ and $0.923$--$0.981$. The analysis uses selected rows. Quadratic-weighted Cohen's $\kappa$ ranges from approximately 0.20--0.30 in Ansible and 0.07--0.49 in GitHub. Across the four rubric judges, annotator-specific correspondence remains heterogeneous. In Ansible, Qwen's pooled correspondence exceeds Gemma by $.172$ [$.014$, $.385$], GPT-OSS-20B by $.300$ [$.088$, $.577$], and GPT-OSS-120B by $.198$ [$.039$, $.415$]. All Qwen contrasts in GitHub, both 120B contrasts with Gemma or GPT-OSS-20B, and the Gemma--20B contrasts include zero.

\subsection{Binned Correspondence}

Table~\ref{tab:pilot_binned} reports the binned correspondence between SpeciTeller scores and pooled pilot labels. Within each corpus, SpeciTeller scores are partitioned into quintiles. Pooled labels are averaged across annotators and linearly normalized to $[0,1]$ for comparability. The pilot uses a small sample and an ordinal scale.

\begin{table}[t]
\caption{Binned correspondence between SpeciTeller scores and pooled pilot labels.}
\label{tab:pilot_binned}
\centering
\small
\setlength{\tabcolsep}{4pt}
\begin{tabular}{lccc}
\toprule
Corpus & Quintile & Mean score & Mean pooled label \\
\midrule
Ansible & Q1 & 0.001 & 0.344 \\
Ansible & Q2 & 0.314 & 0.620 \\
Ansible & Q3 & 0.404 & 0.760 \\
Ansible & Q4 & 0.574 & 0.798 \\
Ansible & Q5 & 1.000 & 0.656 \\
\midrule
GitHub & Q1 & 0.001 & 0.260 \\
GitHub & Q2 & 0.199 & 0.531 \\
GitHub & Q3 & 0.265 & 0.542 \\
GitHub & Q4 & 0.510 & 0.704 \\
GitHub & Q5 & 1.000 & 0.690 \\
\bottomrule
\end{tabular}
\end{table}

\section{Robustness and Surface Proxy Checks}
\label{sec:appendix_robustness_proxy}

Table~\ref{tab:feature_alignment_full} reports the complete visible-cue correlations summarized in Table~\ref{tab:cross_context_profile}.

\begin{table}[t]
\caption{Spearman correlation between SpeciTeller and visible lexical cues.}
\label{tab:feature_alignment_full}
\centering
\setlength{\tabcolsep}{4pt}
\small
\begin{tabular}{lrrrr}
\toprule
Feature & Wiki & GitHub & Ansible & Python \\
\midrule
TFIDF-mean & -0.729 & -0.387 & -0.229 & -0.167 \\
TFIDF-max & -0.682 & -0.289 & -0.176 & -0.092 \\
Tech-ratio & 0.359 & 0.424 & 0.374 & 0.334 \\
Tokens & 0.754 & 0.417 & 0.312 & 0.242 \\
Chars & 0.695 & 0.450 & 0.318 & 0.248 \\
\bottomrule
\end{tabular}
\end{table}

Table~\ref{tab:proxy_decomposition} summarizes a compact decomposition of the aggregate technical-token proxy for the technical corpora. Values are Spearman correlations between predicted specificity and selected deterministic surface indicators. The same indicators were computed uniformly across corpora, but the table omits Wikipedia because it is the general-domain reference corpus rather than technical documentation.

\begin{table}[t]
\caption{Selected technical-token surface proxy decomposition.}
\label{tab:proxy_decomposition}
\centering
\small
\setlength{\tabcolsep}{3pt}
\begin{tabular}{lrrl}
\toprule
Corpus & Tech-ratio & Strongest added & Indicator \\
\midrule
GitHub & 0.424 & 0.392 & identifier \\
Ansible & 0.374 & 0.326 & version/numeric \\
Python Ref & 0.334 & 0.396 & version/numeric \\
\bottomrule
\end{tabular}
\end{table}

The technical corpora do not share one dominant surface signal. GitHub documentation is closest to the aggregate technical-token proxy through identifiers, Ansible shows a mixed surface pattern, and Python reference documentation is more strongly associated with token-shape complexity and version or numeric tokens than with the aggregate proxy. These indicators are observable lexical and token-shape cues. Wikipedia remains dominated in the main analysis by generic expository proxies such as length and TF--IDF.

Table~\ref{tab:normalization_robustness} reports the controlled-edit normalization check. Normalized deltas use each corpus's original score distribution. Directional consistency is identical under raw, z-score, and min--max views.

\begin{table}[t]
\caption{Controlled-edit robustness under within-corpus normalization.}
\label{tab:normalization_robustness}
\centering
\small
\setlength{\tabcolsep}{3pt}
\begin{tabular}{llrrrr}
\toprule
Corpus & Edit & $n$ & Raw mean & $z$ mean & Cons. \\
\midrule
GitHub & add\_spec & 10 & 0.124 & 0.383 & 100\% \\
GitHub & de\_spec & 10 & -0.180 & -0.556 & 100\% \\
GitHub & irrelev. & 10 & -0.001 & -0.004 & 80\% \\
\midrule
Ansible & add\_spec & 8 & 0.054 & 0.178 & 75\% \\
Ansible & de\_spec & 12 & -0.161 & -0.530 & 91.7\% \\
Ansible & irrelev. & 10 & -0.103 & -0.341 & 80\% \\
\bottomrule
\end{tabular}
\end{table}

Within-corpus z-score and min--max normalization are monotonic affine transformations when denominators are positive, so within-corpus Spearman correlations are preserved. Rank associations and controlled-edit directions are stable, while raw corpus mean differences remain learned-scale behavior.

\section{GranuScore Sensitivity and Verification}
\label{sec:granuscore_details}

Table~\ref{tab:cross_context_profile} reports the original-row GranuScore means and rank agreement. Higher native values indicate coarser or more abstract references. Native Wikipedia-minus-technical gaps are $-11.80$ [$-12.14$, $-11.45$], $-11.12$ [$-11.83$, $-10.36$], and $-10.43$ [$-12.05$, $-8.77$]. Strict filtering preserves the ordering and yields gaps of $-11.28$, $-8.08$, and $-6.50$. Excluding rows without extracted referential units also preserves it; no-unit rates are 2.07\%, 0.70\%, 11.57\%, and 0.80\%, and Ansible changes most. Exact-token standardization yields original gaps of $-12.18$, $-11.21$, and $-11.37$, and strict gaps of $-12.03$, $-8.99$, and $-6.99$. Every declared interaction diagnostic is substantial.

The scorer uses official package version 1.0.1 at the repository, model, wheel, official-artifact, and CUDA-image revisions listed in Appendix Table~\ref{tab:component_provenance}. Its scores match the official five examples with maximum absolute error $3.28\times10^{-6}$. Stable identifiers match scores to all 1{,}295{,}205 corpus rows, all 80 pilot rows, and the 60 controlled-edit pairs. The archived configuration, runner, image, score files, bootstrap seeds, and table checksums support reproduction without redistributing sentence text or human labels.

\section{Output-Distribution Diagnostics}
\label{sec:distribution_details}

Table~\ref{tab:predictor_details} reports the comparator repetitions, document-bootstrap intervals, and native-scale corpus gaps summarized in the main text.

\begin{table*}[t]
\caption{Original-row comparison of SpeciTeller (ST) and the pinned post-publication comparator. Adapt mean averages Adapt-1/2/3 by sentence; row SD is the mean sentence-level population SD across repetitions. Intervals resample documents.}
\label{tab:predictor_details}
\centering
\scriptsize
\setlength{\tabcolsep}{3.2pt}
\begin{tabular}{lccccc}
\toprule
Corpus & ST mean & Adapt mean & Adapt-1/2/3 means & Row SD & Spearman [95\% CI] \\
\midrule
Wikipedia & .7534 & .3490 & .3475/.3786/.3208 & .0239 & .5103 [.5008, .5191] \\
GitHub Docs & .4960 & .3993 & .3664/.4418/.3898 & .0324 & .0704 [.0588, .0825] \\
Ansible Docs & .4257 & .3704 & .3857/.3480/.3774 & .0171 & $-.0656$ [$-.1092$, $-.0285$] \\
Python Ref & .2921 & .2825 & .4220/.0680/.3574 & .1542 & .4430 [.4120, .4819] \\
\midrule
\multicolumn{6}{l}{\textit{Wikipedia-minus-technical native-scale gaps}} \\
\midrule
Technical corpus & \multicolumn{2}{c}{ST gap [95\% CI]} & \multicolumn{3}{c}{Adapt mean gap [95\% CI]} \\
GitHub Docs & \multicolumn{2}{c}{+.2574 [.2498, .2655]} & \multicolumn{3}{c}{$-.0504$ [$-.0572$, $-.0444$]} \\
Ansible Docs & \multicolumn{2}{c}{+.3277 [.3090, .3472]} & \multicolumn{3}{c}{$-.0214$ [$-.0321$, $-.0106$]} \\
Python Ref & \multicolumn{2}{c}{+.4613 [.4480, .4774]} & \multicolumn{3}{c}{+.0665 [.0582, .0738]} \\
\bottomrule
\end{tabular}
\end{table*}

Table~\ref{tab:distribution_diagnostics} reports the original-row diagnostics summarized in Section~VI. Effective range is scale-normalized; boundary shares use each source's native theoretical bounds.

\begin{table}[H]
\caption{Selected output-distribution diagnostics. ``Range'' is $(p_{95}-p_{05})/(U-L)$; L5/U5 are percentages in the lower/upper outer 5\%. GS-units removes the no-unit sentinel.}
\label{tab:distribution_diagnostics}
\centering
\scriptsize
\setlength{\tabcolsep}{3.0pt}
\begin{tabular}{llrrr}
\toprule
Source/view & Corpus & Range & L5 & U5 \\
\midrule
SpeciTeller & Wikipedia & .814 & 1.85 & 41.47 \\
SpeciTeller & Python & .938 & 33.27 & 4.59 \\
Adapt-1 & Python & .351 & 0.00 & 0.02 \\
Adapt-2 & Python & .131 & 42.26 & 0.01 \\
Adapt-3 & Python & .360 & 0.00 & 0.02 \\
GranuScore & GitHub & .399 & 0.08 & 1.55 \\
GranuScore & Ansible & .736 & 0.11 & 12.38 \\
GS-units & Ansible & .651 & 0.12 & 0.92 \\
\bottomrule
\end{tabular}
\end{table}

Across all corpora, SpeciTeller effective ranges span .814--.973, while comparator upper-bound mass is at most .024\%. For Python Adapt-2, lower-bound shares are 0.00\%, 42.26\%, and 83.54\% at the outer 1\%, 5\%, and 10\% thresholds. All 18 pre-specified between-predictor comparisons differ in their use of the available scale. The reproducibility supplement provides all 20 model-instance--corpus facets on original and strict rows, native quantiles, boundary shares, tie and mass-point diagnostics, and the full 20-facet histogram.

\section{Corpus and Component Provenance}
\label{sec:repro_details}

\begingroup
\newcommand{\ProvValue}[2]{\nolinkurl{#2}}

\begin{table}[H]
\caption{Corpus source provenance. Supporting reproducibility files are in the SpecTech repository.}
\label{tab:corpus_provenance}
\centering
\scriptsize
\setlength{\tabcolsep}{2.6pt}
\begin{tabular}{p{0.23\columnwidth}p{0.39\columnwidth}p{0.28\columnwidth}}
\toprule
Corpus & Identifier type & Pinned identifier \\
\midrule
GitHub Docs & Source Git commit & \ProvValue{github_source_git}{5b8a8c96f169} \\
Ansible Docs & Source Git commit & \ProvValue{ansible_source_git}{83bbdd618bb6} \\
Python Ref & Source archive SHA-256 & \ProvValue{python_source_sha}{9e98419a01ef} \\
Wikipedia & Source archive SHA-256 & \ProvValue{wiki_source_sha}{5b08ca86a17f} \\
Wikipedia & Deterministic build & 2{,}000 files \\
\bottomrule
\end{tabular}
\end{table}

\begin{table}[H]
\caption{Scorer and model provenance. Supporting reproducibility files are in the SpecTech repository.}
\label{tab:component_provenance}
\centering
\scriptsize
\setlength{\tabcolsep}{2.2pt}
\begin{tabular}{p{0.22\columnwidth}p{0.35\columnwidth}p{0.34\columnwidth}}
\toprule
Component & Identifier type & Pinned identifier \\
\midrule
SpeciTeller & Repository Git commit & \ProvValue{speciteller_git}{218cb5a389b3} \\
SpeciTeller & Released-data SHA-256 & \ProvValue{speciteller_data_sha}{dd70443a77b6} \\
SpeciTeller & LIBLINEAR Git commit & \ProvValue{liblinear_git}{491c9f1188b9} \\
SpeciTeller & Container image tag & \ProvValue{speciteller_image_tag}{speciteller:py27} \\
Ko comparator & Upstream Git commit & \ProvValue{ko_upstream_git}{36f8e835e9dc} \\
Ko comparator & Container image tag & \ProvValue{ko_image_tag}{spectech-ko-specificity:36f8e835-official-release-v1} \\
Ko comparator & GloVe text SHA-256 & \ProvValue{ko_glove_sha}{9641ef45628f} \\
GranuScore & Repository Git commit & \ProvValue{granuscore_git}{73438cdbb06a} \\
GranuScore & Model Git revision & \ProvValue{granuscore_model_git}{b170cbfa5bb7} \\
GranuScore & Package-wheel SHA-256 & \ProvValue{granuscore_wheel_sha}{8d0325512243} \\
GranuScore & Container image digest & \ProvValue{granuscore_image_digest}{d637a2aebb39} \\
GranuScore & Official LGB artifact SHA-256 & \ProvValue{granuscore_lgb_sha}{5aba55b9e13c} \\
Qwen3-14B & Ollama list ID & \ProvValue{qwen_list_id}{bdbd181c33f2} \\
Qwen3-14B & Model-blob SHA-256 & \ProvValue{qwen_blob_sha}{a8cc1361f314} \\
Gemma4-12B & Ollama list ID & \ProvValue{gemma_list_id}{4eb23ef187e2} \\
Gemma4-12B & Ollama manifest SHA-256 & \ProvValue{gemma_manifest_sha}{4eb23ef187e2} \\
Gemma4-12B & Model-blob SHA-256 & \ProvValue{gemma_blob_sha}{1278394b6936} \\
Gemma4-12B & Projector-blob SHA-256 & \ProvValue{gemma_projector_sha}{675ad6e68101} \\
GPT-OSS-20B & Ollama list ID & \ProvValue{gptoss20_list_id}{17052f91a42e} \\
GPT-OSS-20B & Ollama manifest SHA-256 & \ProvValue{gptoss20_manifest_sha}{17052f91a42e} \\
GPT-OSS-20B & Model-blob SHA-256 & \ProvValue{gptoss20_blob_sha}{e7b273f96360} \\
GPT-OSS-120B & Ollama manifest SHA-256 & \ProvValue{gptoss120_manifest_sha}{a951a23b46a1} \\
GPT-OSS-120B & Model-blob SHA-256 & \ProvValue{gptoss120_blob_sha}{6be6d66a3f54} \\
\bottomrule
\end{tabular}
\end{table}

\begin{table}[H]
\caption{Protocol and evidence provenance. Supporting reproducibility files are in the SpecTech repository.}
\label{tab:evidence_provenance}
\centering
\scriptsize
\setlength{\tabcolsep}{2.2pt}
\begin{tabular}{p{0.25\columnwidth}p{0.38\columnwidth}p{0.28\columnwidth}}
\toprule
Component & Identifier type & Pinned identifier \\
\midrule
Generator prompt G & Config-file SHA-256 & \ProvValue{generator_prompt_sha}{c14fc73dbf12} \\
Rubric prompt R & Config-file SHA-256 & \ProvValue{rubric_prompt_sha}{41686f9c2a98} \\
Ko released-domain diagnostic & Config-file SHA-256 & \ProvValue{ko_released_diagnostic_sha}{9ea05db72f47} \\
Gemma review & Blinded review file SHA-256 & \ProvValue{gemma_workbook_sha}{73312e07427d} \\
GPT-OSS-120B review & Blinded review file SHA-256 & \ProvValue{gptoss120_workbook_sha}{5c84775834cf} \\
DGX run archive & Archive SHA-256 & \ProvValue{dgx_return_sha}{bf6f5d9b5564} \\
DGX run archive & Run ID & \ProvValue{dgx_run_id}{dgx-5411bb15be9f} \\
\bottomrule
\end{tabular}
\end{table}

\endgroup

We adapted the Ko comparator three times to each of the four corpora, reporting Adapt-1, Adapt-2, and Adapt-3 separately and their sentence-wise mean as Adapt mean. This implementation uses a different objective from the 2019 experimental artifact; the 2019 checkpoint was not released, the published Twitter results were not reproduced, and augmentation is unseeded.

\textit{Released-domain diagnostic.} The Twitter evaluation used the released 50{,}000-tweet adaptation set and 984 annotated rows; its Spearman, Kendall, and MAE were .4686, .3219, and .1454, versus published values of $.676\pm.004$, $.487\pm.005$, and $.113\pm.001$. The supplemental diagnostic changed only the released target-domain mapping and produced 844 Yelp and 919 Movie predictions per run. Across three runs, Yelp yielded $.6142\pm.0140$, $.4442\pm.0124$, and $.1388\pm.0056$, versus $.750\pm.016$, $.555\pm.010$, and $.107\pm.003$; Movie yielded $.2616\pm.0268$, $.1839\pm.0188$, and $.1642\pm.0009$, versus $.702\pm.003$, $.519\pm.015$, and $.114\pm.001$. Observed variability is population SD across three runs; the published SD convention is not stated. These results evaluate the pinned implementation, which is used as a separate comparator.

Table~\ref{tab:component_provenance} lists the model identities used for the 80-sentence pilot. Qwen and the remote GPT-OSS-120B run use Ollama 0.32.5; the matched local Gemma and GPT-OSS-20B replications use Ollama 0.32.9. All four have 80/80 coverage and use identical inputs and rubric, one blind sentence per fresh conversation, a strict score schema, and 10{,}000 paired bootstraps. The DGX run archive has an accepted provenance waiver: missing attempt, post-load identity, and receipt-chronology records are audit-completeness gaps, not evidence of tampering.

Both generators used the same 60 source sentences and produced three rewrites per source without seeing scores. All 180 GPT-OSS-120B candidates received blinded review, and the comparison used 20{,}000 source-level bootstrap replicates. GranuScore 1.0.1 reproduces the official examples and scores every corpus sentence, pilot sentence, and controlled edit.

\subsection{Candidate-Selection Example}
\label{sec:traceable_candidate_example}

\begingroup
\scriptsize
The Gemma case with case-ID prefix \texttt{5a510bd8} is one of the 40 cases in which SpeciTeller and the first-candidate baseline select different candidates.

The neutral-rewrite source is ``When it is Inactive, this link will no longer be valid.'' \textit{Candidate 1} is ``This link will no longer be valid when it is Inactive.'' ($\Delta=-0.005299$). \textit{Candidate 2} is ``Once it becomes Inactive, this link will no longer be valid.'' ($\Delta=+0.004479$). \textit{Candidate 3} is ``Once it is Inactive, this link will no longer be valid.'' ($\Delta=+0.000401$).

All three received blinded rating 3 (equal specificity, neutral/direction-valid). The first-candidate baseline selects Candidate 1; SpeciTeller selects Candidate 3 because its absolute change is smallest. The case is illustrative, not representative or additional aggregate evidence.
\endgroup

\section{Detailed Sensitivity Boundaries}
\label{sec:detailed_boundaries}

GranuScore's no-unit rates are 0.70\%--11.57\%; unit-only sensitivity preserves ordering but changes Ansible most. Controlled-edit specificity can co-vary with length, lexical realization, modality, or proposition content. The strict condition retains 35.6\%--70.9\% of rows.

Common support retains 83.0\%/77.0\% of original/strict Wikipedia rows; Python has the weakest overlap (effective-sample-size fractions 0.559/0.622; maximum weights 8.241/6.493). Normalization and proxy decomposition preserve ranks and edit directions.

\textit{Candidate-selection diagnostics.} For SpeciTeller on Gemma, the gain concentrates in add-specific cases: +33.3 points over 18 paired sources [95\% CI +11.1, +55.6], compared with zero for 20 neutral cases. Adapt-2 and Adapt-3 are each 10 points below the first-candidate baseline on neutral cases [$-30$, +10]. Removing the session's first 30 ratings preserves the estimated directions with less precision: SpeciTeller is +8.9 points over 45 sources [$-2.2$, +20.0] and GranuScore is +9.1 over 44 [$-2.3$, +20.5]; every comparator interval includes zero.

Across all candidates, GPT-OSS-120B direction-validity is 51.1\% versus Gemma's 76.1\%, a paired difference of $-25.0$ points [$-35.0$, $-15.0$]. Within GPT-OSS-120B, direction-validity is 51.9\% for add-specific, 71.2\% for de-specific, and 28.3\% for neutral candidates. The between-set difference remains $-25.3$ and $-26.4$ points after excluding each session's first 20 or 30 ratings, while score-3 prevalence is 23.9\% versus 25.6\%. One evaluator's separate sessions leave candidate-set and session effects inseparable.

Within GPT-OSS-120B, score deltas have positive human association: SpeciTeller $\rho=.570$ [.419, .692], Adapt-1/2/3 $.317$--$.332$ (intervals .117--.529), and aligned GranuScore $.268$ [.070, .451]. Nevertheless, SpeciTeller selects 56.7\% direction-valid candidates versus 51.7\% for the first-candidate baseline, a +5.0-point gain [$-6.7$, +16.7], and is adverse for neutral cases (20.0\% versus 25.0\%). The secondary consensus reaches 63.3\%, a +11.7-point gain [+1.7, +21.7]. All primary single-score GPT-OSS-120B intervals and between-set gain intervals include zero.

For Gemma, the counterfactual surface proxy would accept 98 valid and 12 invalid edits and reject 39 valid and 31 invalid edits (27.9\% false-accept; 28.5\% false-reject). For GPT-OSS-120B, the same counterfactual counts would be 43, 21, 49, and 67 (23.9\% false-accept; 53.3\% false-reject [39.6\%, 66.7\%]). The paired false-rejection difference is +24.8 points [+9.7, +39.4].

\section{LLM Roles and Prompt Templates}
\label{sec:llm_prompt_roles}

Table~\ref{tab:llm_prompt_roles} distinguishes generation, rubric judging, and fixed-text scoring. Generators receive source sentences and edit requests; rubric judges receive one pilot sentence per call; scoring follows generation.

\begin{table}[H]
\caption{Experimental roles and inputs. Rubric-judge results concern the 80-sentence pilot, not either generated candidate set; a model name appearing in both rows does not imply self-judging.}
\label{tab:llm_prompt_roles}
\centering
\scriptsize
\setlength{\tabcolsep}{2pt}
\begin{tabular}{p{0.24\columnwidth}p{0.13\columnwidth}p{0.25\columnwidth}p{0.15\columnwidth}c}
\toprule
Model(s) & Role & Visible input & Output & ID \\
\midrule
\texttt{gemma4:12b}; \texttt{gpt-oss:120b} & Candidate generator & Original sentence plus one add, reduce, or neutral request & One edited sentence & G \\
\texttt{qwen3:14b}; \texttt{gemma4:12b}; \texttt{gpt-oss:20b}; \texttt{gpt-oss:120b} & Rubric judge & One pilot sentence & One integer, 1--5 & R \\
SpeciTeller; Adapt-1/2/3; GranuScore & Post-generation scorer & Fixed candidate text & Scalar score & --- \\
\bottomrule
\end{tabular}
\end{table}

\begingroup
\newcommand{\promptlinebreak}{\par\noindent}
\newcommand{\promptparagraphbreak}{\par\smallskip\noindent}

\subsection{Exact Generator Prompt G}

Gemma4-12B and GPT-OSS-120B use the same prompt and input format below, with different model and runtime settings. Braced text is a literal substitution field.

\textit{System prompt.}
\begin{quote}
\footnotesize\ttfamily\raggedright
Generate one controlled rewrite for a sentence-specificity study. Preserve the sentence's main meaning, factual commitments, polarity, modality, and communicative intent. Make only the requested kind of change. Do not assume hidden context or invent unsupported facts. Return exactly one grammatical sentence in the required JSON field and no explanation.
\end{quote}

\textit{Add-specific user template.}
\begin{quote}
\footnotesize\ttfamily\raggedright
Rewrite the original sentence so it is more specific while preserving its meaning and factual correctness.\promptparagraphbreak
Original sentence:\promptlinebreak
\{sentence\_original\}
\end{quote}

\textit{De-specify user template.}
\begin{quote}
\footnotesize\ttfamily\raggedright
Rewrite the original sentence so it is less specific while preserving its meaning and factual correctness.\promptparagraphbreak
Original sentence:\promptlinebreak
\{sentence\_original\}
\end{quote}

\textit{Neutral-rewrite user template.}
\begin{quote}
\footnotesize\ttfamily\raggedright
Paraphrase the original sentence while keeping approximately the same specificity, meaning, and factual correctness.\promptparagraphbreak
Original sentence:\promptlinebreak
\{sentence\_original\}
\end{quote}

The interface required exactly one JSON object with schema \texttt{\{ ``edited\_sentence'': string \}}, containing one sentence and no explanation. These prompts requested meaning and factual preservation; that request was not treated as evidence that either property held. The pre-specified blindness rule was:

\begin{quote}
\footnotesize\ttfamily\raggedright
No corpus, identifiers, author edit, prior model output, prior attempt, proxy, predictor score, rubric score, human label, or review information enters a request.
\end{quote}

\subsection{Exact Rubric-Judge Prompt R}

All four judges used the same rubric and response format.

\textit{System prompt.}
\begin{quote}
\footnotesize\ttfamily\raggedright
Assign one sentence-specificity score using this fixed five-point rubric. 1 = very vague or very general; 2 = somewhat vague; 3 = moderately specific; 4 = specific; 5 = very specific or highly concrete. A sentence is more specific when it mentions concrete details such as versions, file paths, commands, parameters, identifiers, numbers, identifiable entities, or explicit operational information. A sentence is less specific when it uses general language, makes high-level statements without details, or lacks concrete identifiers. Use your first impression; there are no right or wrong answers. As a generic illustration, 'Restart the service' is likely lower specificity, while 'Restart the nginx service using systemctl restart nginx on host web-03' is likely higher specificity. Judge only the supplied sentence without assuming surrounding context. Return only a JSON object with the integer field score and no explanation.
\end{quote}

\textit{User template.}
\begin{quote}
\footnotesize\ttfamily\raggedright
Sentence:\promptlinebreak
\{sent\_text\}
\end{quote}

The response schema was exactly one field, \texttt{\{ ``score'': integer \}}, with the integer restricted to 1--5 and no additional fields. Each call contained one pilot sentence in a fresh conversation and received one attempt. Deterministic settings were temperature 0, top-$k$ 1, top-$p$ 1, repeat penalty 1, seed 20260809, and context length 4096; Qwen and Gemma allowed 16 output tokens, while GPT-OSS allowed 128 for its interface-only reasoning/final budget. The system prompt contained no pilot example, human label, model or predictor score, bucket, corpus identity, or numeric worked input--output example. These judges scored the same 80-sentence pilot; they did not score, select, or evaluate either generated candidate set.

\subsection{Role Boundary}

Candidate generation was score-blind; predictor and GranuScore values were computed after candidates were fixed. Rubric judging used separate pilot sentences. Although Gemma4-12B and GPT-OSS-120B served both roles, separate runs and inputs prevent self-judging or self-selection.

\endgroup

\end{document}